\documentclass[pdflatex,sn-mathphys-num]{sn-jnl}%

\usepackage{graphicx}%
\usepackage{multirow}%
\usepackage{amsmath,amssymb,amsfonts}%
\usepackage{amsthm}%
\usepackage{mathrsfs}%
\usepackage[title]{appendix}%
\usepackage{xcolor}%
\usepackage{textcomp}%
\usepackage{manyfoot}%
\usepackage{booktabs}%
\usepackage{algorithm}%
\usepackage{algorithmicx}%
\usepackage{algpseudocode}%
\usepackage{listings}%

\graphicspath{ {./images/} }

\newtheorem{mydef}{Theorem}
\theoremstyle{thmstyletwo}%

\begin{document}

\title[Article Title]{SOAR: Social and Opinion-Aware Recommendations}


\author*[1]{\fnm{Giannis} \sur{Christoforidis}}\email{icchrist@csd.auth.gr}

\author[1]{\fnm{Apostolos} \spfx{N.} \sur{Papadopoulos}}\email{papadopo@csd.auth.gr}

\affil*[1]{\orgdiv{School of Informatics}, \orgname{Aristotle University of Thessaloniki}, \orgaddress{\street{AUTH Campus}, \city{Thessaloniki}, \postcode{54124}, \country{Greece}}}

\abstract{One of the enduring challenges in online platforms is recommendation. Both communities and advertisers face difficulties in analyzing complex user behavior, identifying habits and evolving preferences, and providing timely recommendations that are relevant to user interests. In particular, point-of-interest (POI) recommendation poses the additional challenge of identifying the factors that influence user behavior, capturing user tendencies, and generating accurate, personalized recommendations for each individual.
In this work, we propose SOAR, a Social and Opinion-Aware Recommendation model that incorporates both the user’s friendship network and textual reviews to assess whether a location is suitable for a specific user, with a particular emphasis on the sentiment of shared reviews. Experimental results in real-world datasets demonstrate that incorporating reviews into the recommendation process improves the precision of the recommendation.}

\keywords{Point-of-Interest Recommendation, Social Network Analysis, Sentiment-Aware Recommender Systems}

\maketitle

\section{Introduction}
\label{sec:intro}

Social networks and, in general, online presence have increased dramatically in recent years. People frequently log their activities on the Internet and use social networks to share opinions, thoughts, and other personal experiences. Due to the increase in traffic on such networks, there have been a plethora of research opportunities regarding analyzing the user's data in order to understand their tendencies and preferences and thus be able to predict/suggest future personalized actions. A challenging approach has always been POI recommendation, especially with the many factors that contribute to the user's final choice of location.

Traditional recommendation approaches often rely solely on user-location visitation, overlooking the complex social and contextual factors that influence users' preferences and idiosyncrasies. A user can be influenced by a number of factors, including social, temporal, geographical, and other. A challenge lies in detecting these aspects and also applying them to each user specifically to achieve personalized recommendations.

A wide variety of parameters can affect a user's preference, but ultimately the most indicative result of the user's visitation is the final review. The language behind each review can indicate not only the sentiment of the user, but also their mentality and thought process, which, in turn, can reveal several factors that the user weighs more when choosing both a location to visit and to avoid. Using the text in POI reviews, we can formulate a semantic network by aligning the common factors detected in the text and combining them with the friendship network. As a result, we can pinpoint each users' important positive and negative factor and provide suggestions on POIs with similar mentality of the specific user. It is crucial to understand and point out that every user can find something appeasing where another user finds it appalling - a characteristic that can be very challenging.

\begin{figure}[!t]
    \centering
    \includegraphics[width=\linewidth]{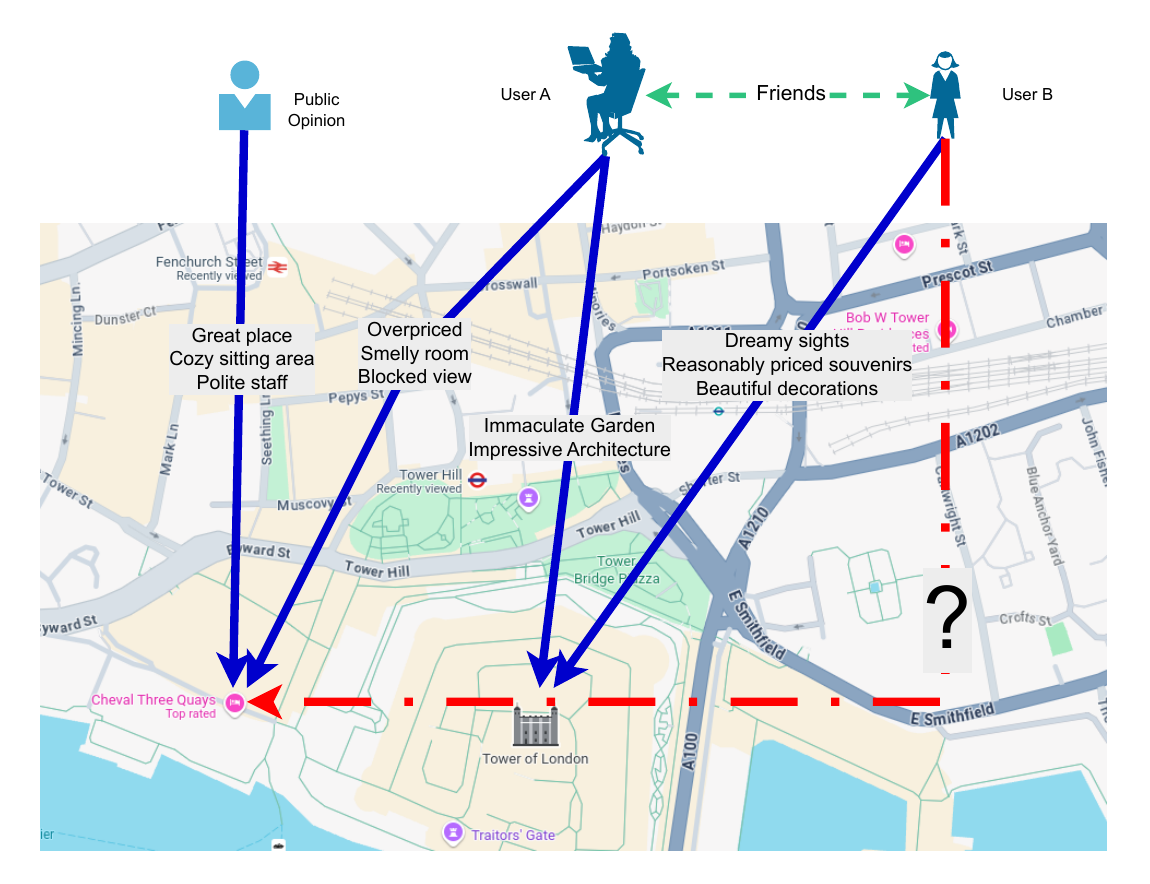}
    \caption{Recommendation Example}
    \label{fig:example}
\end{figure}

Figure~\ref{fig:example} illustrates a simple example of a recommendation challenge that can be encountered in the real world. The silhouette on the left depicts the public/dominant opinion over a hotel in London - it's a top rated location with many visitors and positive reviews. However, we observe some negative traits in the review of User A, and the friend of this user, User B, considers visiting that location. As they also share many common preferences such as the quality of the view and the prices, it is unlikely that the user will prefer that location despite its general good fame.

In this paper, we present a Social and Opinion-Aware Recommendation (SOAR) model that integrates text-based reviews with user network information and historical activity to generate personalized recommendations. Our goal is to evaluate all locations based on their historic activity, but more importantly to identify each user's idiosyncrasies by analyzing the user's friend's reviews, and extracting the traits they find either appealing or appalling. While it is important to take account the public opinion on popular locations we also weight heavily upon the user's peculiarities.

More specifically, we employ SBERT~\cite{SBERT} to encode textual reviews into semantic embeddings. This allows us to encapture the different traits a user may like or dislike, and present them in a vectorized form. Additionally, we apply a FAISS~\cite{FAISS} clustering process that groups related reviews — and, consequently, associated locations—based on shared characteristics. This process yields a graph structure that represents connections between locations that share similar characteristics or traits. Additionally, we calculate the sentiment value of each review with Hugging Face's pipeline in an effort to detect each user's idiosyncrasies from the user's group of friends. We then incorporate this graph with the user's social network and historical check-in data, generating unified embeddings through RandNE\cite{RandNE}.

Finally, to improve personalization, we adjust each embedding by weighting reviews from friends of the user more heavily in the recommendation process. To achieve that, we assign the initial score on each location based on the public sentiment and then we value both negative and positive reviews on the user's friends specifically. The score of the affected locations is then altered based on the sentiment of the user's friends' reviews, adding or subtracting points accordingly to the ratio of their friends' mentality, and as such each suggestion is now made personalized for the specific user.

The contribution of our work is summarized as follows: 
\begin{itemize}
    \item \textit{Sentiment Value}: We split reviews into positive and negative based on their context. We then insert them into a graph along with the friends so the can determine the final score of each location not only by its total reviews, but by the average reaction from the friends of each user.

    \item \textit{Semantic Similarity}: Each review has a different point of view and is greatly inspired by the mentality of the user. It is important to contextualize the review and denote the important factors that characterize the location. By doing so, we can determine if a user prefers to visit this location based on those factors regardless of the sentiment of the review author, as each person has their own preferences.

    \item \textit{Personalized Recommendations}: Each person has their own preferences and a circle of friends that influence the user's decision-making process. Our algorithm integrates both general sentiment and volume of review for each location, as well as opinions of the user’s friends, since public opinion and peer sentiment may differ.
\end{itemize}

The rest of the work is organized as follows. Related work is presented in Section~\ref{sec:related}. The methodology and technical details of the proposed approach are given in Section~\ref{sec:proposed}. Performance evaluation results are presented and discussed in Section~\ref{sec:experiments}, while Section~\ref{sec:conclusions} concludes the work and briefly discusses future research directions.

\section{Related Work}
\label{sec:related}


In this section, we review prior research in the areas of both POI recommendation and text analysis.

A considerable body of recent research in this field has focused on POI recommendation based on users’ historical interactions with their environment. Some works \cite{Meng2024,Liu2024,Xu2021} engage into the social aspect of the users, aiming to establish their preferences based on their friendship network and other social connections. Other models \cite{Gan2023,Fu2024} focus on the geographical cues to improve recommendation accuracy, often including spatio-temporal information to include long-term and short term interest as well as seasonal data. In addition, researchers \cite{Zhang2024,Fang2023} have also attempted to combine the social and spatio-temporal geographical information in order to incorporate more data-points and thus achieve better personalized recommendations. While it is important to include a variety of information, we should not overlook the textual data such as reviews, as they are more informative and precise to the users' preference than any other data.

In an effort to capture both global and local contextual features along with POI attributes, Liu et al.~\cite{Liu2023} proposed KDRank, a model designed to identify general popularity patterns as well as localized contextual information based on users’ friendship network. While distinguishing the influence of friends from dominant behavioral trends is crucial, their approach employs a variation of the PageRank algorithm~\cite{Bianchini2005} to weight the relative importance of each friend in the recommendation process. However, this method inherently limits the contribution of less influential friends and, in the absence of textual or semantic information, may overlook valuable contextual cues. In our work, we address this limitation by modeling the distinct behavioral traits of each user and integrating the friendship network adaptively to adjust the final ranking scores of candidate locations.

We very often see multiple variations of BERT~\cite{BERT} being used throughout the research field~\cite{Acharya2025,Ko2022} in the text/review analysis section.

Similarly to our work, Huang et al.~\cite{Huang23} uses BERT-CL, a clustering modification of the BERT algorithm, which focuses on embedding the representation of the text and also classifies positive and negative reviews into a score-based system. While on the same path, it completely ignores every other parameter, such as the friendship network and only calculates the future public opinion based on historic data and as such does not personalize each recommendation.

SBERT was also used in MMPOI, a model developed by Xu et al.~\cite{Xu2024}, which also incorporates text to categorize locations. More specifically, they use Meta-data, reviews and transform images into descriptions to further contextualize the location. Further, they also include geographic and multi-modal trajectory path, a technique we've also seen in other works\cite{Christoforidis2021,Li2024,Qin2023} which also utilize sequences in POI visitation. While incorporating many aspects of a user's preferences they don't seem to align user's with their review's characteristic and identify the traits each user, or group of users, are attracted to.

Wang et al.~\cite{Wang23} proposed a model called STaTRL, which finds aspects within the review text and characterizes each as positive or negative in an effort to determine the boons and banes of each POI. Whereas there are similarities in finding aspects in each review, they do not take into account the friendship network and depend heavily on the "popular" opinion. Reviews tend to be heavily subjective opinions and as such valuing the opinions of ones friends outweigh the dominant consensus. By linking the review text's aspects and connecting them to the users we can identify the user's tendencies and as such determine the ones that contribute positively or negatively regardless of the generic "popular" mentality.

An et. al.~\cite{An2022} focused on sentiment-based tourist spot recommendation using a CNN-LSTM model to infer ratings from textual reviews and incorporate environmental context: weather, season, and spatial proximity. Initially, a TF-IDF mechanism is applied to eliminate words that are overly dependent on specific tourist spots, as identified through morpheme analysis. CNN extracts the pooling layer that is concatenated by LSTM for sentence encoding. Although effective for general context-aware recommendations, their approach does not model user-specific preferences or relationships between users. In contrast, our method builds on the idea of exploiting review text but further integrates user-network information and personalized embeddings to provide individualized POI recommendations.

Contractor et al.~\cite{Contractor2021} address the problem of answering natural-language questions about POIs by leveraging tourism reviews. Their work frames POI retrieval as a question-answering task that combines textual and spatial reasoning, employing a multi-stage pipeline that clusters review sentences, retrieves relevant POIs, and subsequently re-ranks them based on semantic relevance to the query. In contrast, our approach, instead of processing queries in isolation, incorporates user-specific information, such as social network connections and historical interactions, alongside review semantics. By encoding review texts and constructing a graph that links semantically related reviews and locations, our model captures both contextual and relational dependencies among users and POIs. This graph-based representation enables personalized and interpretable recommendations, whereas their framework primarily emphasizes general question understanding and retrieval effectiveness.

Yao et al.~\cite{Yao2025} propose a sequential POI recommendation framework that explicitly leverages user reviews and feedback by first applying a two-level clustering over review text and then modeling user trajectories with an LSTM to capture sequential visitation patterns and user similarity. Their method emphasizes sequential modeling of POI discovery (using recurrent networks) and incorporates multifaceted contextual features to mitigate data sparsity and cold-start problems via learned user similarity from feedback. Instead of engaging sequential pattern learning and feedback-driven similarity, we instead identify each user's personal idiosyncrasies by connecting him to the patterns in their friends' reviews by utilizing the friendship network - something that is missing from their model, yet we consider essential for providing personalized recommendations.

While our work leverages user reviews, it differs from prior review-aware recommendation approaches in its primary focus. Existing methods largely concentrate on extracting rich semantic representations from review text and optimizing text modeling techniques. In contrast, our approach emphasizes the integration of review-derived signals with the underlying friendship network, aiming to capture the interplay between textual preferences and social structure. As such, our contribution lies in multi-source fusion rather than advances in review text modeling itself. To ensure a fair evaluation, we compare our method against variants that utilize review information without combining it with the friendship network, thereby isolating the impact of the social component. Consequently, direct comparisons with methods that primarily target sophisticated review representation learning are not entirely appropriate, as they address a different aspect of the recommendation problem.

\section{\bfseries{\sc SOAR}: The Proposed Approach}
\label{sec:proposed}

In this section we first present the fundamentals of our research and then describe in detail our proposed methodology and algorithm for the data analysis and personalized recommendation system. Table \ref{tab:Symbols and Notations} depicts the most commonly used symbols.

\subsection{Fundamentals}

\begin{table}[!ht]
    \centering
    \caption{Frequently Used Symbols}
    \begin{tabular}{ccccc}
    \toprule
    Symbol & Description\\
    \midrule
    $\mathcal{U}, \mathcal{L}, \mathcal{R}, \mathcal{C}$ & Set of Users, $\mathcal{U}=\{u_1,...,u_n\}$, \\
                & set of locations $\mathcal{L}=\{l_1,...,l_m\}$, 		\\ 
			    & set of Reviews $\mathcal{R}=\{r_1,...,r_k\}$, 	\\
			    & set of Check-ins $\mathcal{C}=\{c_1,...,c_o\}$ 			\\
    $\mathcal{F}_{u_i}$ & set of Users' Friends $\mathcal{F}=\{f{u_1},...,f{u_n}\}$, 	\\
    $c_{u_i}$		& user's check-in								\\
    $\mathbf{U}^\mathcal{R}, \mathbf{U}^\mathcal{G}$		& Embeddings of Reviews and Graphs respectively	\\
    & $\mathbf{U}^\mathcal{R}=(u_1^\mathcal{R},...,u_{d_r}^\mathcal{R})$ \\
    & $\mathbf{U}^\mathcal{G}=(u_1^\mathcal{G},...,u_{d_g}^\mathcal{G})$ \\
    \bottomrule
    \end{tabular}
    \label{tab:Symbols and Notations}
\end{table}

\begin{mydef}{POI:}\label{def:POI} is a unique location where at least one user has checked-in. Every POI is represented as a tuple: $<l_{id}$, longitude, latitude$>$.
\end{mydef}

\begin{mydef}{Check-in:}\label{def:Check-in} is a visit from User $u$ to a Location $l$ with review $r$ made after the visit forming a tuple $c_i=<u,l,r>$. A user can have many check-ins to the same place.
\end{mydef}

\begin{mydef}{User-User graph:\label{def:uu_graph}}
is a unipartite undirected graph representing the friendship relations among users in the social network, denoted as $G_{uu} = (\mathcal{U}, \mathcal{E})$, where $\mathcal{U}$ is the set of users and $\mathcal{E} \subseteq \mathcal{U} \times \mathcal{U}$ is the set of edges representing friendship relations.
\end{mydef}

\begin{mydef}{Review-Review graph:\label{def:rr_graph}}
is a unipartite undirected graph representing similarity relationships among reviews, denoted as $G_{rr} = (\mathcal{R}, \mathcal{E})$, where $\mathcal{R}$ is the set of reviews and $\mathcal{E} \subseteq \mathcal{R} \times \mathcal{R}$ is the set of edges connecting similar reviews. Each review $r_i \in \mathcal{R}$ is represented by an embedding vector:
\begin{equation}
\mathbf{u}_i^{\mathcal{R}} \in \mathbb{R}^d
\label{eq:rr1}
\end{equation}
obtained from a pretrained language model. The edge set $\mathcal{E}$ is constructed based on nearest-neighbor relationships in the embedding space. Specifically, for each $r_i \in \mathcal{R}$, we define edges to its $k$ nearest neighbors:
\begin{equation}
\mathcal{E} = \bigcup_{i=1}^{|\mathcal{R}|} \left\{ \{ r_i, r_j \} \;\middle|\; r_j \in \text{NN}_k\bigl(\mathbf{u}_i^{\mathcal{R}}\bigr) \right\}.
\end{equation}
where $\text{NN}_k(\cdot)$ denotes the set of $k$ nearest neighbors in the embedding space.
\end{mydef}

\begin{mydef}{User-Location graph:\label{def:ul_graph}} is a bipartite undirected consisting of connections based on the check-ins. It is denoted as $G_{ul}=\mathcal{U}\cup \mathcal{L}$ where $\mathcal{L}$ is the set of Locations and $\mathcal{U}$ is the set of Users, and $\mathcal{C} \subseteq \mathcal{U} \times \mathcal{L}$ the set of check-ins. We define a bipartite undirected graph

\begin{equation}
G_{ul} = (\mathcal{U} \cup \mathcal{L}, \mathcal{E}),
\end{equation}

where 

\begin{equation}
\mathcal{E} = \bigl\{ \{u, \ell\} \;\big|\; (u, \ell) \in \mathcal{C},\; u \in \mathcal{U},\; \ell \in \mathcal{L} \bigr\}.
\end{equation}
\end{mydef}

\subsection{Sentiment Analysis of Reviews}
\label{subsec:reviews}

To extract the polarity of user reviews, we employed a transformer-based sentiment classifier using the \textit{Hugging Face} implementation of the \texttt{DistilBERT} model. This model performs binary classification, assigning each review either a positive or negative sentiment label.

Formally, for each review $r_i \in \mathcal{R}$, we compute its sentiment as follows:

\begin{equation}
s_i = f_{\text{BERT}}(r_i)
\end{equation}

where $f_{\text{BERT}}$ denotes the DistilBERT-based sentiment function.  
The model outputs two logits corresponding to the probability of each class (positive and negative):

\begin{equation}
\mathbf{p}_i = \text{softmax}(\mathbf{z}_i)
\quad \text{where} \quad
\mathbf{z}_i = \text{BERT}(r_i)
\end{equation}

with 
\begin{equation}
\mathbf{p}_i = [p_i^{(\text{neg})}, p_i^{(\text{pos})}]
\quad \text{such that} \quad
p_i^{(c)} = \frac{e^{z_i^{(c)}}}{e^{z_i^{(\text{pos})}} + e^{z_i^{(\text{neg})}}}
\end{equation}

The final sentiment label is determined by:

\begin{equation}
y_i =
\begin{cases}
1, & \text{if } p_i^{(\text{pos})} \ge p_i^{(\text{neg})} \\
0, & \text{otherwise}
\end{cases}
\end{equation}

where $y_i = 1$ denotes a positive sentiment and $y_i = 0$ denotes a negative sentiment.  

\subsection{Proposed Approach}

In this section, we explain in detail the functionality of our algorithm. We start by describing the processing of the data available from the dataset, then the method of forming the embeddings, and finally the process we use to calculate the final personalized recommendations.

First, we form the graphs as defined in Theorems~\ref{def:uu_graph}, \ref{def:rr_graph}, and \ref{def:ul_graph}. More specifically, for the reviews we used Sentence-BERT\cite{SBERT}, a modification of the BERT \cite{BERT} architecture optimized for sentence-level embeddings as depicted in Figure~\ref{fig:BERT}. Each review $r=(w_1, w_2, \cdots,w_n)$ is tokenized and mapped to the input embeddings by:

\begin{figure}[!b]
    \centering
    \includegraphics[width=\linewidth]{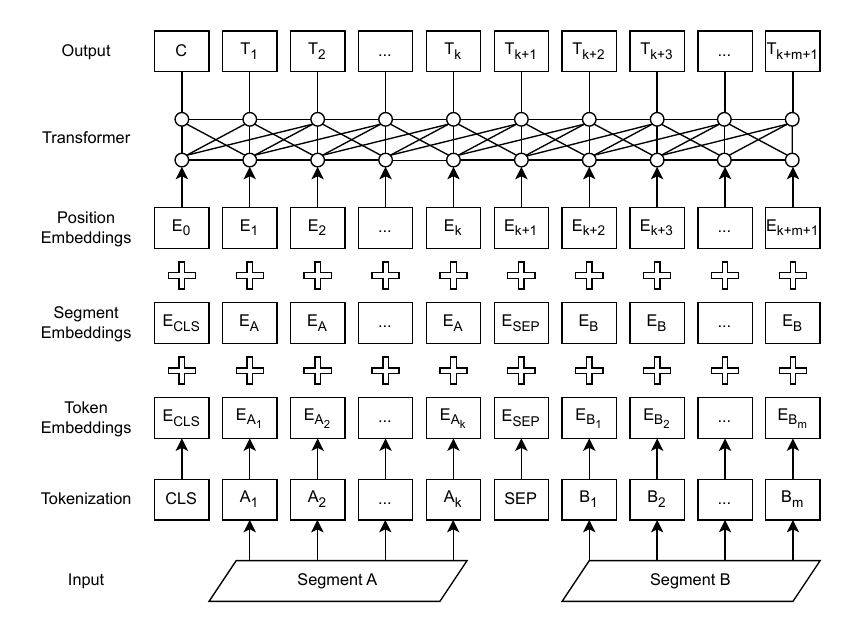}
    \caption{Bert example.}
    \label{fig:BERT}
\end{figure}

\begin{equation}
\mathbf{x}_i = \text{Embed}(w_i) + \text{PosEnc}(i)
\end{equation}

Where $\text{Embed}(\cdot)$ is the word embedding function and $\text{PosEnc}(i)$ denotes the positional encoding. The resulting token embeddings $(\mathbf{x}_1,\cdots,\mathbf{x}_n)$ are processed by a transformer encoder consisting of multiple self-attention layers, producing contextualized token representations

\begin{equation}
\mathbf{H} = (\mathbf{h}_1, \dots, \mathbf{h}_n) = \text{Transformer}(\mathbf{x}_1, \dots, \mathbf{x}_n), \quad \mathbf{H} \in \mathbb{R}^{n \times d}
\end{equation}

To derive a fixed-size sentence embedding, SBERT applies a pooling operation over the sequence of token embeddings. In our implementation, we used mean pooling, defined as

\begin{equation}
\mathbf{s} = \frac{1}{n} \sum_{i=1}^{n} \mathbf{h}_i
\end{equation}

where $s=\in \mathbb{R}^d$ represents the embedding of the entire review text. The similarity between two reviews $r_a$ and $r_b$ is then computed using cosine similarity:

\begin{equation}
\text{sim}(\mathbf{u}_a^\mathcal{R}, \mathbf{u}_b^\mathcal{R}) = \frac{\mathbf{u}_a^\mathcal{R} \cdot \mathbf{u}_b^\mathcal{R}}{\|\mathbf{u}_a^\mathcal{R}\| \, \|\mathbf{u}_b^\mathcal{R}\|}
\end{equation}

\begin{figure}[!b]
    \centering
    \includegraphics[width=\linewidth]{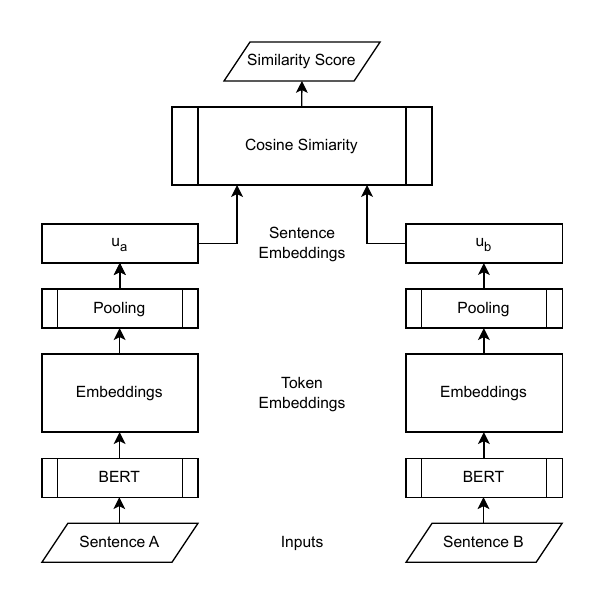}
    \caption{Similarity Score Computation.}
    \label{fig:SimiarityBERT}
\end{figure}

SBERT itself is fine-tuned using a Siamese network structure with objectives such as semantic textual similarity regression, natural language inference classification, or triplet loss ranking as illustrated in Figure~\ref{fig:SimiarityBERT}. For example, in the STS setting, given a gold similarity score $y\in [0,5]$

\begin{equation}
\mathcal{L} = \left( y - \text{sim}(\mathbf{u}_a^\mathcal{R}, \mathbf{u}_b^\mathcal{R}) \right)^2
\end{equation}

Algorithm~\ref{alg:review_sentiment} describes the procedure to produce the Review Graph $G^\mathcal{R}$ and the Sentiment of the reviews. As described in previous sections, we use SBERT to produce the initial review embeddings, then we use FAISS to connect nearby neighbors and finally we apply Hugging Face pipeline to produce the final sentiment rating that characterize a review as positive or negative.

\begin{algorithm}
    \caption{Review Graph and Sentiment Analysis}
    \begin{algorithmic}[1]
    \label{alg:review_sentiment}

    \Require{Reviews: $\mathcal{R}$, Sentence Transformer Model $m^b$, Pipeline Model $m^p$, Review radius $range$}

    \Ensure{Review Graph $G^\mathcal{R}$, Review Sentiment $\mathcal{S}$}

        \State $\mathbf{U}^\mathcal{R}\gets m^b.encode(\mathcal{R})$ \Comment{Generate Review Embeddings}

        \State $map\gets faiss.index(\mathbf{U}^\mathcal{R})$ \Comment{Generate Review Graph}
        \For{$r$ in $\mathcal{R}$}
            \State $G^\mathcal{R}.add(<r,faiss.neighbors(range)>)$
        \EndFor

        \State $pipeline.initialize(m^p)$ \Comment{Sentiment Analysis}
        
        \For{$r$ in $\mathcal{R}$}
            \If{$pipeline.classify(r)>0.5$}
                \State $\mathcal{S}_r\gets positive$
             \Else
                \State $\mathcal{S}_r\gets negative$
            \EndIf      
        \EndFor
    \State \Return $\mathbf{U}$
    \end{algorithmic}
\end{algorithm}

With the 3 graphs formed, we re-identify each node and combine them to one unipartite graph. Then, we use the RandNE\cite{RandNE} to calculate the final embeddings. More specifically we first form the adjacency matrix of $\mathcal{G}$, $
\mathbf{A} \in \mathbb{R}^{N \times N}$ and preserve the high-order proximity by considering powers of $\mathbf{A}$:

\begin{equation}
\mathbf{S} = \Bigl( w_0 \mathbf{I} + w_1 \mathbf{A} + w_2 \mathbf{A}^2 + \dots + w_q \mathbf{A}^q \Bigr)
\end{equation}

where $\mathbf{I}$ is the identity matrix and $w_i$ are the pre-defined weights controlling the contribution of $k$-hop neighbors. This step ensures that nodes connected indirectly through multiple hops influence each other’s embedding. The final embeddings are computed via iterative random projection:

\begin{equation}
\mathbf{\mathbf{U}^\mathcal{G}} = \mathbf{S} \cdot \mathbf{R}
\end{equation}

where $\mathbf{R}$ is a random Gaussian matrix $R_{ij}\sim\mathcal{N}(0,1)$. Algorithm \ref{alg:embeddings} describes the process of calculating the embeddings using Gaussian Random Matrix with iterative random projections.

\begin{algorithm}
    \caption{Calculation of the Embeddings}
    \label{alg:embeddings}
    \begin{algorithmic}[1]
    \Require{User, POIs and User-Review Graphs: $G_u, G_l, G_{ul}$, Embedding dimension $d$, k-hop weights $w$}

    \Ensure{Embeddings $\mathbf{\mathbf{U}}$}

        \State $\mathbf{G} \gets combine(G_u, G_l, G_{ul})$
        \State $N\gets count\_nodes( \mathbf{G})$
        \State $\mathbf{A}$ $\gets sparse\_matrix(\mathbf{G})$ 
        \State $\mathbf{A} \gets \mathbf{A}+\mathbf{A}^T$ \Comment{Calculate the Adjacency matrix}
        
        \State $\mathbf{U}_{list}$ $\gets Gaussian\_Random\_Matrix(N\times d)$ 
        \For{$i$ in $w$}
            \State $\mathbf{U}_{list}.append(\mathbf{A} \cdot \mathbf{U}_{list}[-1])$ \Comment{Initialize the Gaussian Random Matrix and random projection}
        \EndFor

        \State $\mathbf{U}\gets empty\_array(d)$
        \For{$i$ in $w$}
            \State $\mathbf{U}\gets w_i\times \mathbf{U}_{list}[w_i]$ \Comment{Calculate the embeddings}
        \EndFor
    \State \Return $\mathbf{U}$
    \end{algorithmic}
\end{algorithm}

\subsection{Personalized Recommendations}
\label{subsec:personalized_rec}

For a given future check-in event marked as a tuple $c=<u_i,l_j>$ where user $u_i\in\mathcal{U}$ visits location $l_j\in\mathcal{L}$ we first retrieve all locations within a radius $r$ around $l_j$:

\begin{equation}
\mathcal{N}_r(\ell_j) = \bigl\{ \ell_t \in \mathcal{L} \;\big|\; \text{dist}(\ell_t, \ell_j) \le r \bigr\}.
\end{equation}

Each location $l_t\in\mathcal{N}_r(\ell_j)$ is assigned a friend-review weight $w_{i,t}\in[0,1]$, defined as the proportion of positive reviews among the user's friends who have visited that location and alters its score:

\begin{equation}
w_{i,t} = 
\begin{cases}
\dfrac{P_{i,t}}{T_{i,t}} \cdot \sigma(s_i), & \text{if } T_{i,t} > 0, \\[1.0em]
0, & \text{otherwise.}
\end{cases}
\end{equation}

Where $P_{i,t}$ and $T_{i,t}$ denote the number of positive and total reviews left by the friends of user $u_i$ for location $l_t$. 

Algorithm \ref{alg:recommendations} describes the process we use for each location to calculate the personalized score by modifying the embeddings according to each user's friendship network. For every location we alter its score based on the user's reviews and value them more than the dominant popular opinion.

We continue to compute the personalized similarity score for each location $l_t$:

\begin{equation}
\mathbf{s}_{i,t} = \mathbf{U}_i^{\mathcal{G}} \cdot 
\bigl( \mathbf{U}_{l_t} \bigr)^{\!\top} 
+ \mathbf{w_{i,t}}
\end{equation}

Finally, we rank all candidate locations in descending order according to their relevance scores $s_{i,t}$. The location with the highest score is considered the most suitable recommendation for user $u_i$. Moreover, the model supports Top-$N$ recommendation by selecting the $N$ highest-ranked locations, thus providing a ranked list of recommendations.

\begin{algorithm}
    \caption{Producing Recommendations}
    \label{alg:recommendations}
    \begin{algorithmic}[1]
    \Require{Graph Embeddings: $\mathbf{U}$, Friendship Network: $\mathcal{F}$, User's Reviews: $\mathcal{R}$, User: $u$, User's Location: $lat,lon$}

    \Ensure{Final Score}

        \State $NL$ $\gets$ find\_nearby($lat,lon$)
        \State $Modified\_U \gets \mathbf{U}(NL)$ \Comment{Find Nearby Locations and Modify the Embeddings}

        \State $final\_scores \gets \mathbf{U}(u) \cdot Modified\_U^T$ \Comment{}{Inner Product}
        \State $std\gets standard\_deviation(final\_scores)$ \Comment{Calculate the final scores}
        \For{$l$ in $NL$}
            \State $positive\_reviews$, $total\_reviews$ $\gets$ count\_reviews\_of\_friends($u$)
            
            \If{$total\_reviews$ $equals$ 0}
                \State {$location\_score$ $\gets$ $0.5$}
            \Else
                \State {$location\_score$ $\gets$ $\frac{positive\_reviews}{total\_reviews}$}
            \EndIf
            \State $final\_scores(l) \gets final\_scores(l)+ std*location\_score$
        \EndFor
    \State \Return $final\_scores$
    \end{algorithmic}
\end{algorithm}

\subsection{Example}

\begin{figure}[!t]
    \centering
    \includegraphics[width=\linewidth]{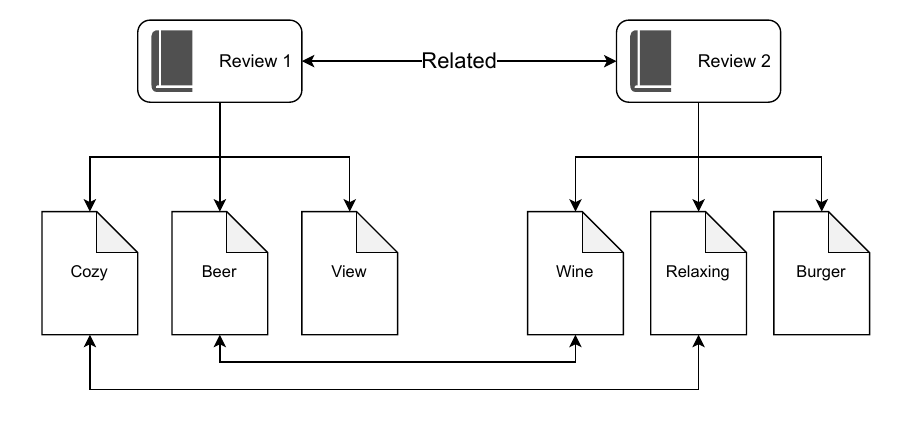}
    \caption{Example of Review Connection.}
    \label{fig:example_review_graph}
\end{figure}

In this section, we describe how our algorithm works by providing some examples that describe its key points. Starting with the review handling, we try to connect locations with each other by detecting and connecting their characteristics. Figure~\ref{fig:example_review_graph} shows the relation between 2 reviews of 2 different locations.

\begin{figure}
        \centering
        \includegraphics{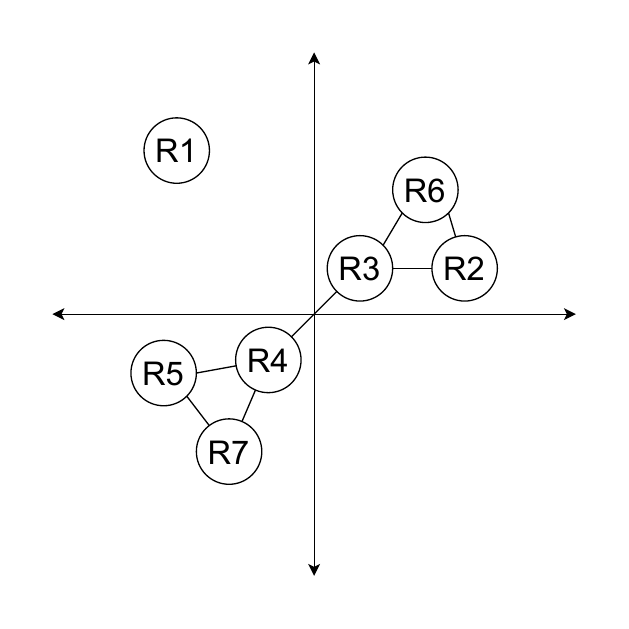}
        \caption{Example of Review Connection in Vector Space.}
        \label{fig:example_review_vector}
    \end{figure}

Although reviews may contain overlapping vocabulary, our analysis goes beyond lexical similarity by generating semantic embeddings for each review. The reviews are then linked based on their proximity in the embedding space, enabling us to capture deeper semantic relationships, as illustrated in Figure~\ref{fig:example_review_vector}.

\begin{figure}
        \centering
        \includegraphics[width=0.9\textwidth]{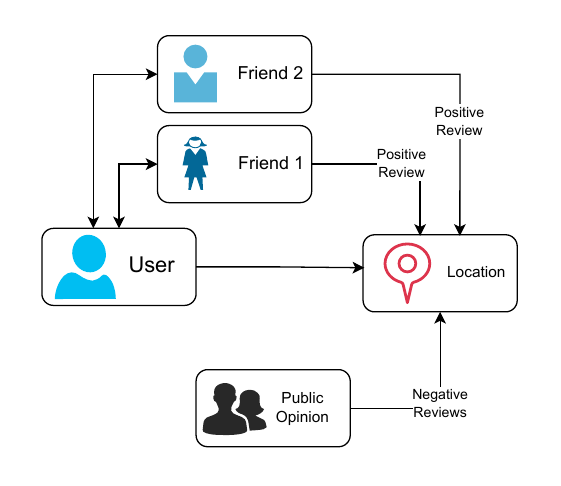}
        \caption{Example of Friend's influence.}
        \label{fig:example_peer}
    \end{figure}

To personalize the recommendation process we take into account the user's friends' reviews. Figure~\ref{fig:example_peer} shows an example of a user's friends ranking highly a specific location whereas the common opinion differ. It is not uncommon for a small group of people to value a characteristic of a location that many people avoid. In this example, since we value the user's opinions more than the general consensus, we will value the specific location higher than normal in spite of the overwhelming negative reviews from the rest of the users.

\section{Performance Evaluation}
\label{sec:experiments}
The source code associated with the proposed methodology is publicly available at \url{https://github.com/thedx4/soar}.
\subsection{Dataset}

We use the yelp dataset \footnote{https://www.kaggle.com/datasets/yelp-dataset/yelp-dataset} that contains approximately 7 million reviews, with an average of 105 words per review and 10 friends per user, as shown in Table~\ref{tab:dataset}. Furthermore, the data set includes 150,000 locations in the United States, of which 100,000 have at least 100 neighboring locations within a 1-kilometer radius.

\begin{table}[!ht]
    \centering
    \caption{{\sc Yelp Dataset} statistical data.}
    \begin{tabular}{ccccc}
    \toprule
    Statistic & Value\\
    \midrule
    Total Number of Reviews & 6,990,280 \\
    Total Number of Words & 732,415,837 \\
    Total Number of Locations & 150,346 \\
    Average Number of Words & 104.78 \\
    Median Review Length (words) & 75 \\
    Std Deviation of Review Length (words) & 97.92 \\
    Longest Review (words) & 1070 \\
    Average Number of Friends & 10.02 \\
    \bottomrule
    \end{tabular}
    \label{tab:dataset}
\end{table}

\subsection{Review Data}

The review graph is generated following the procedure outlined in Theorem~\ref{def:rr_graph}. We employ the pretrained model "all-MiniLM-L6-v2" with an embedding dimension of 384, which offers a favorable trade-off between computational cost and embedding quality, ensuring efficient yet expressive representations of textual reviews.

To identify the nearest neighbors we used Faiss, and to determine an appropriate search radius $r$ for the range-based neighbor retrieval, we conducted an empirical analysis across multiple candidate values. For each radius, we performed a range search over a random sample of queries and recorded descriptive statistics regarding the number of neighbors retrieved per query (minimum, maximum, average, and extreme counts).

The radius was selected such that the average number of neighbors per query fell within a predefined target range ensuring sufficient neighborhood density, while simultaneously avoiding excessively large neighborhoods. Table~\ref{tab:radius_stats} shows the average number of neighbors along with the maximum number of neighbors and the number of nodes that exceed 200 neighbors. As we observer from the graph, higher values result in exponentially higher connections making the graph more dense.

\begin{table}[!ht]
\centering
\caption{Review statistics for different radius values}
\setlength{\tabcolsep}{8pt} 
\renewcommand{\arraystretch}{1.2} 
\begin{tabular}{cccccc}
\hline
\textbf{Radius} & \textbf{Avg} & \textbf{Max} & \textbf{\textgreater200} \\
\hline
0.500 & 1.83 & 153 & 0 \\
0.525 & 2.94 & 330 & 2 \\
0.550 & 5.58 & 630 & 5 \\
0.575 & 11.30 & 1166 & 12 \\
0.600 & 23.91 & 2248 & 26 \\
\textbf{0.625} & \textbf{51.03} & \textbf{4336} & \textbf{49}\\
0.650 & 108.99 & 7677 & 88 \\
0.675 & 229.26 & 12623 & 141 \\
0.700 & 476.20 & 19231 & 232  \\
\hline
\end{tabular}
\label{tab:radius_stats}
\end{table}

We selected the balanced radius of $0.625$ with limiting the number of connections to 50 closest neighbors so that it preserves the local structure of the embedding space without over-saturating the similarity graph with weak or distant connections. The final review graph network result an average number of $7.66$ connections with the most tight cluster exceeding $50$ connections and its standard deviation is $14.8$. As such, we avoid over-clustering tight review groups while maintaining enough edges to characterize them.

To determine the polarity (positive or negative) of each review, we employed a pretrained transformer-based sentiment analysis model from the Hugging Face library. Specifically, we used the "distilbert-base-uncased-finetuned-sst-2-english model", which provides a binary classification between positive and negative sentiment. The model was accessed through the transformers’ pipeline("sentiment-analysis") interface, which performs tokenization, inference, and softmax scoring. Each review's sentiment ultimately contributes to determine the user's group of friends' mentality over a specific network and weight them against the popular opinion as described in section~\ref{subsec:personalized_rec}.

\subsection{Evaluation Metrics}

To evaluate the performance of the recommendation model, we employ standard classification metrics derived from the confusion matrix components: True Positives (TP), False Positives (FP), False Negatives (FN), True Negatives (TN), and Top-K hits. They are defined as follows:

\begin{itemize}
    \item \textbf{True Positives (TP):} The number of instances where we suggest the location for the user, and the user had a positive sentiment.
    \item \textbf{False Positives (FP):} The number of instances where we suggest the location for the user, and the user had a negative sentiment.
    \item \textbf{False Negatives (FN):} The number of instances where we did not suggest the location for the user, and the user had a positive sentiment.
    \item \textbf{True Negatives (TN):} The number of instances where we did not suggest the location for the user, and the user had a negative sentiment.
    \item \textbf{Top-K hits:} The number of instances where we suggest the location for the user, and the user visited the location regardless of the sentiment.
\end{itemize}

Using these quantities, the evaluation metrics are calculated using the following equations:

\begin{align}
    \text{Precision} &= \frac{TP}{TP + FP} \label{eq:precision} \\[6pt]
    \text{Recall} &= \frac{TP}{TP + FN} \label{eq:recall} \\[6pt]
    \text{F1-score} &= 2 \times \frac{\text{Precision} \times \text{Recall}}{\text{Precision} + \text{Recall}} \label{eq:f1} \\[6pt]
    \text{Accuracy} &= \frac{TP + TN}{TP + FP + FN + TN} \label{eq:accuracy} \\[6pt]
    \text{Top-K hits} &= \frac{TP + FP}{TP + FP + FN + TN} \label{eq:top-k}
\end{align}

Precision reflects the proportion of correctly identified positive instances among all predicted positives, while Recall measures the ability to identify all relevant instances. The F1-score provides a harmonic mean of Precision and Recall, balancing both metrics. Finally, Accuracy quantifies the overall correctness of the model’s predictions. Finally, the top-K hits indicate the accuracy of the algorithm while ignoring the final sentiment - which indicates the effectiveness of the algorithm to predict ultimately the user's final choice.

\subsection{Experimental Results}

For our experiments, we used an embedding dimension of 128, a spatial search radius of 400 meters, and a training split comprising 80\% of the available data. To further assess the contribution of our approach, we compared performance on a subset of test cases where users were located within the vicinity of businesses reviewed by their friends.

\begin{figure}
        \centering
        \includegraphics[width=0.5\textwidth]{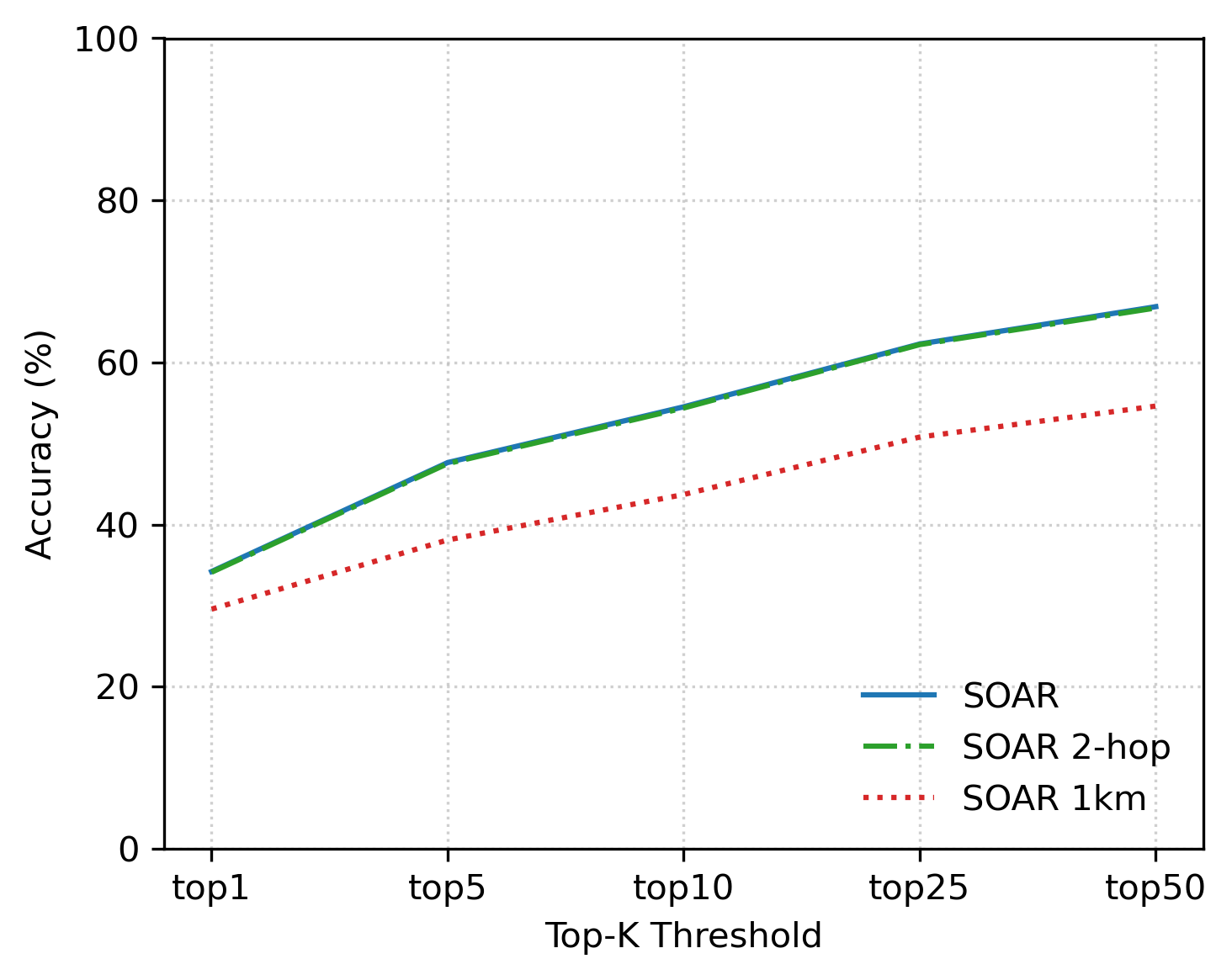}
        \caption{Accuracy relative to radius}
        \label{fig:accuracy_3models}
    \end{figure}

\begin{figure}
        \centering
        \includegraphics[width=0.5\textwidth]{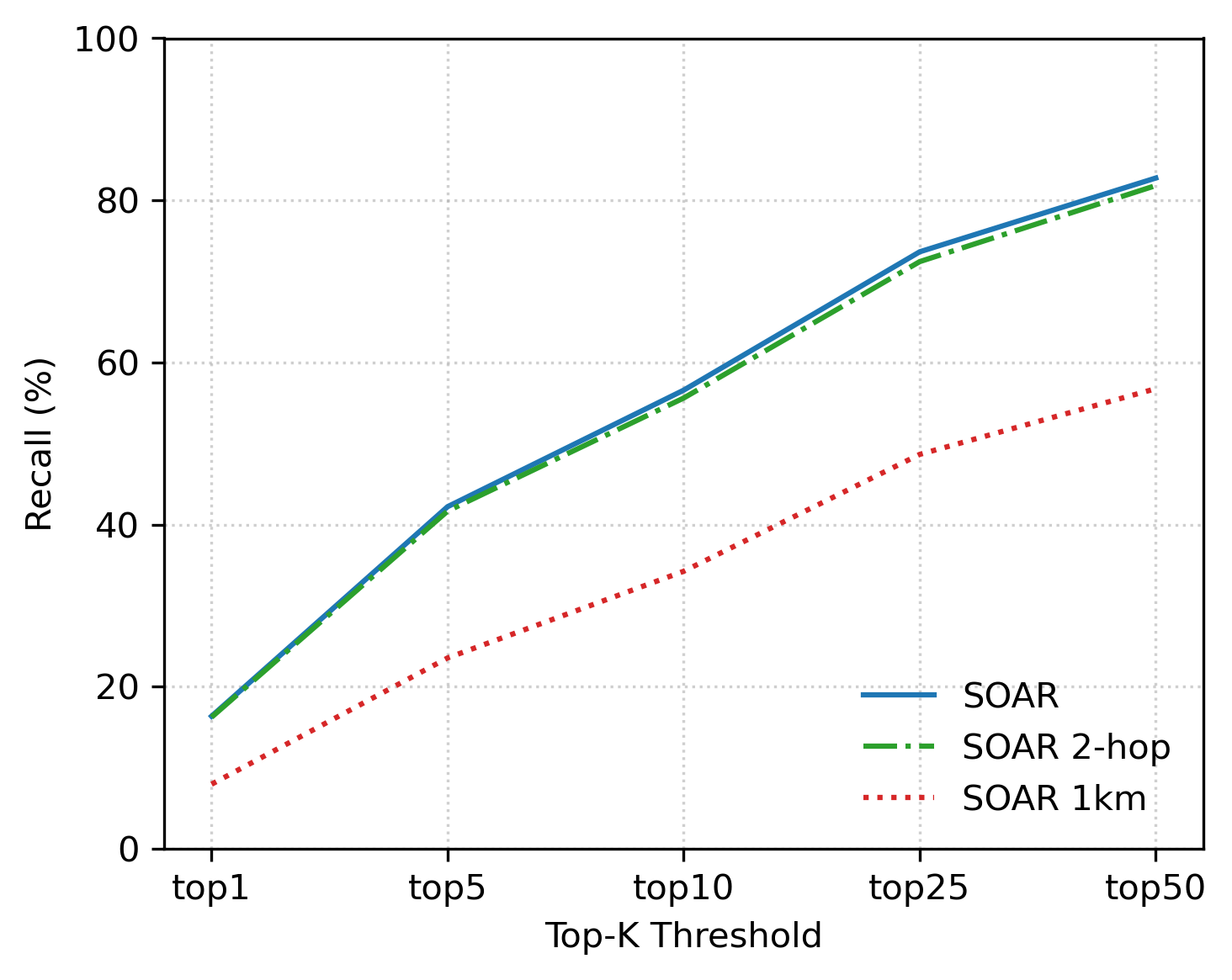}
        \caption{Recall relative to radius}
        \label{fig:recall_3models}
    \end{figure}

\begin{figure}
        \centering
        \includegraphics[width=0.5\textwidth]{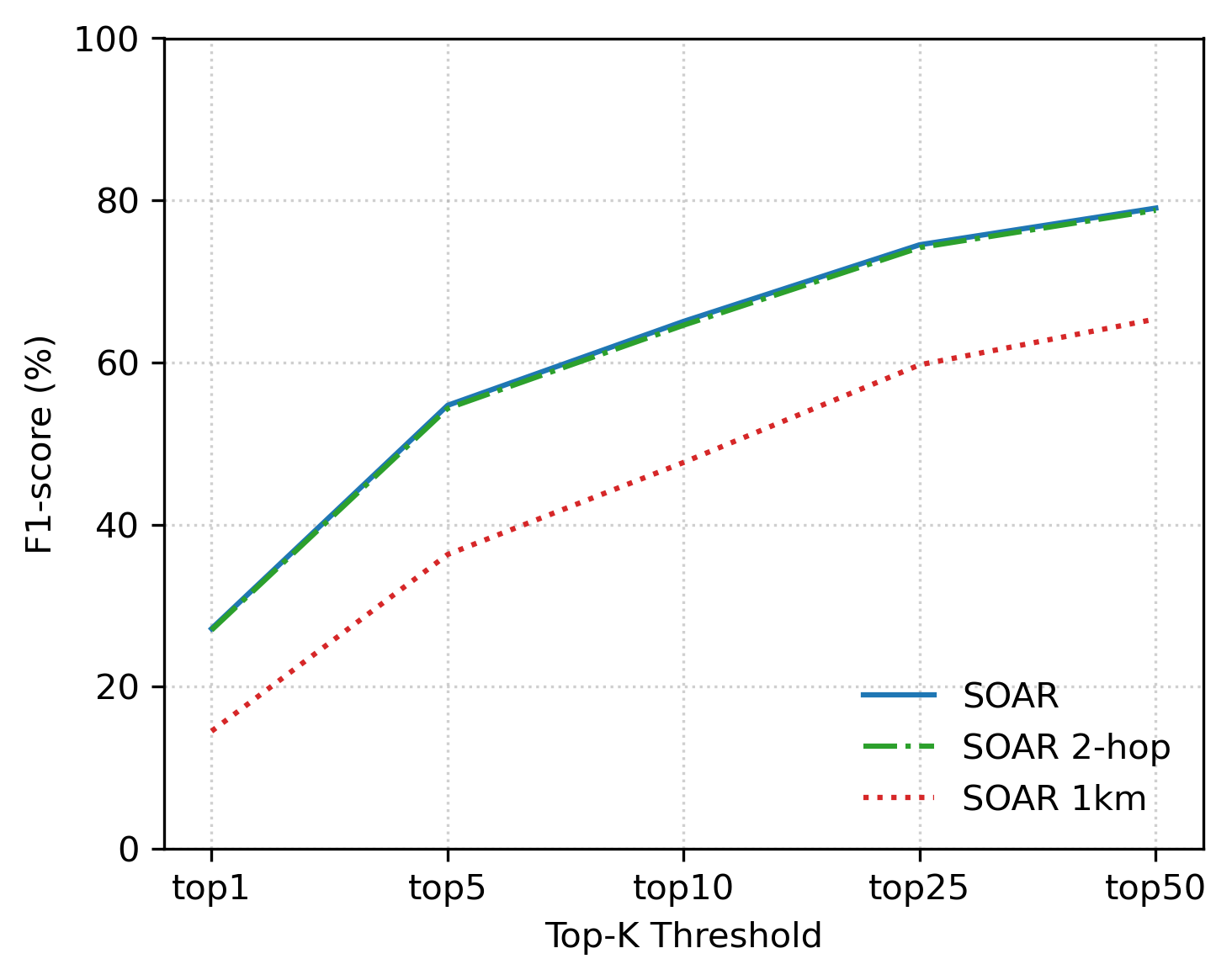}
        \caption{F1-score relative to radius}
        \label{fig:f1_3models}
    \end{figure}

\begin{figure}
        \centering
        \includegraphics[width=0.5\textwidth]{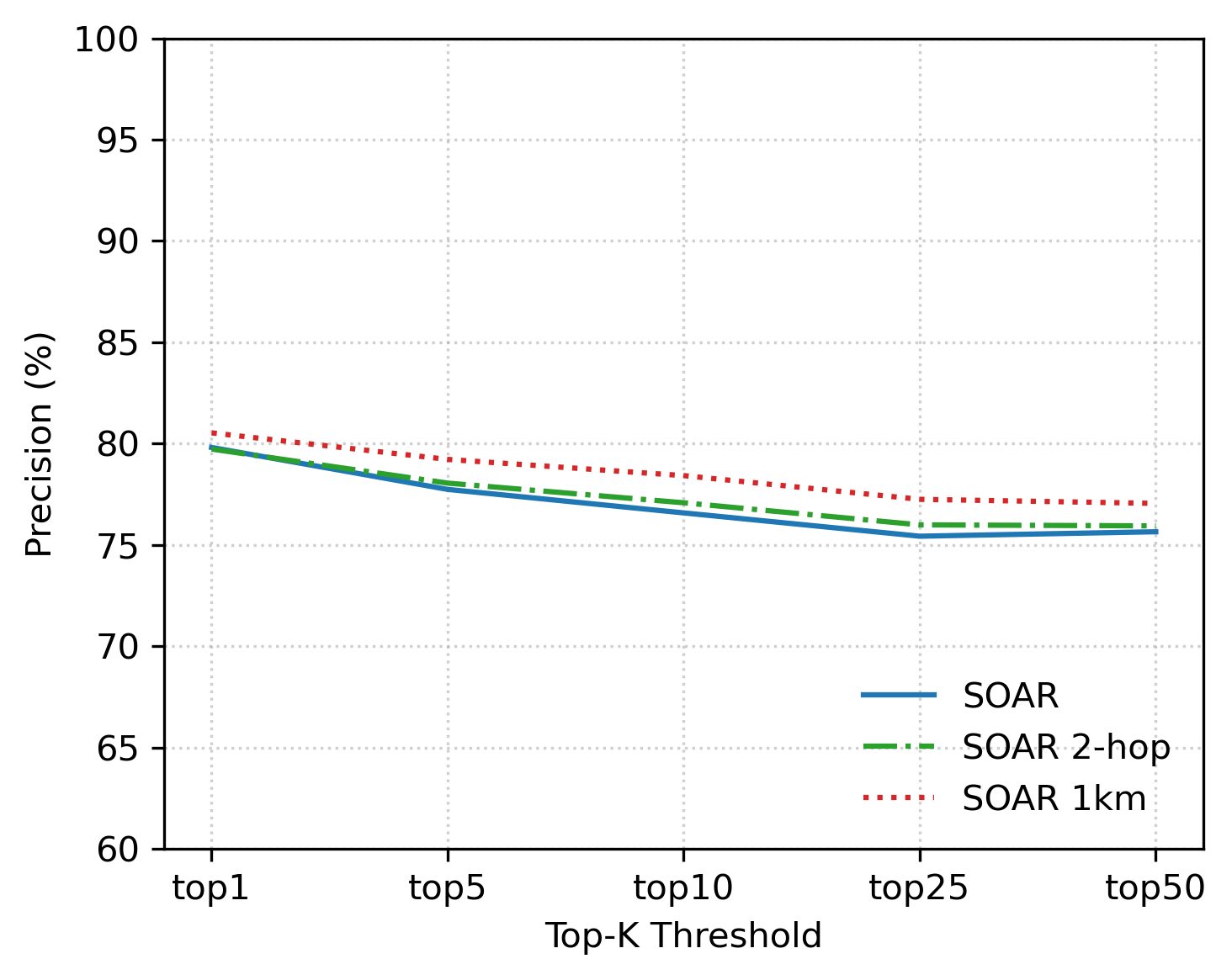}
        \caption{Precision relative to radius}
        \label{fig:Precision_3models}
    \end{figure}

\begin{figure}
        \centering
        \includegraphics[width=0.5\textwidth]{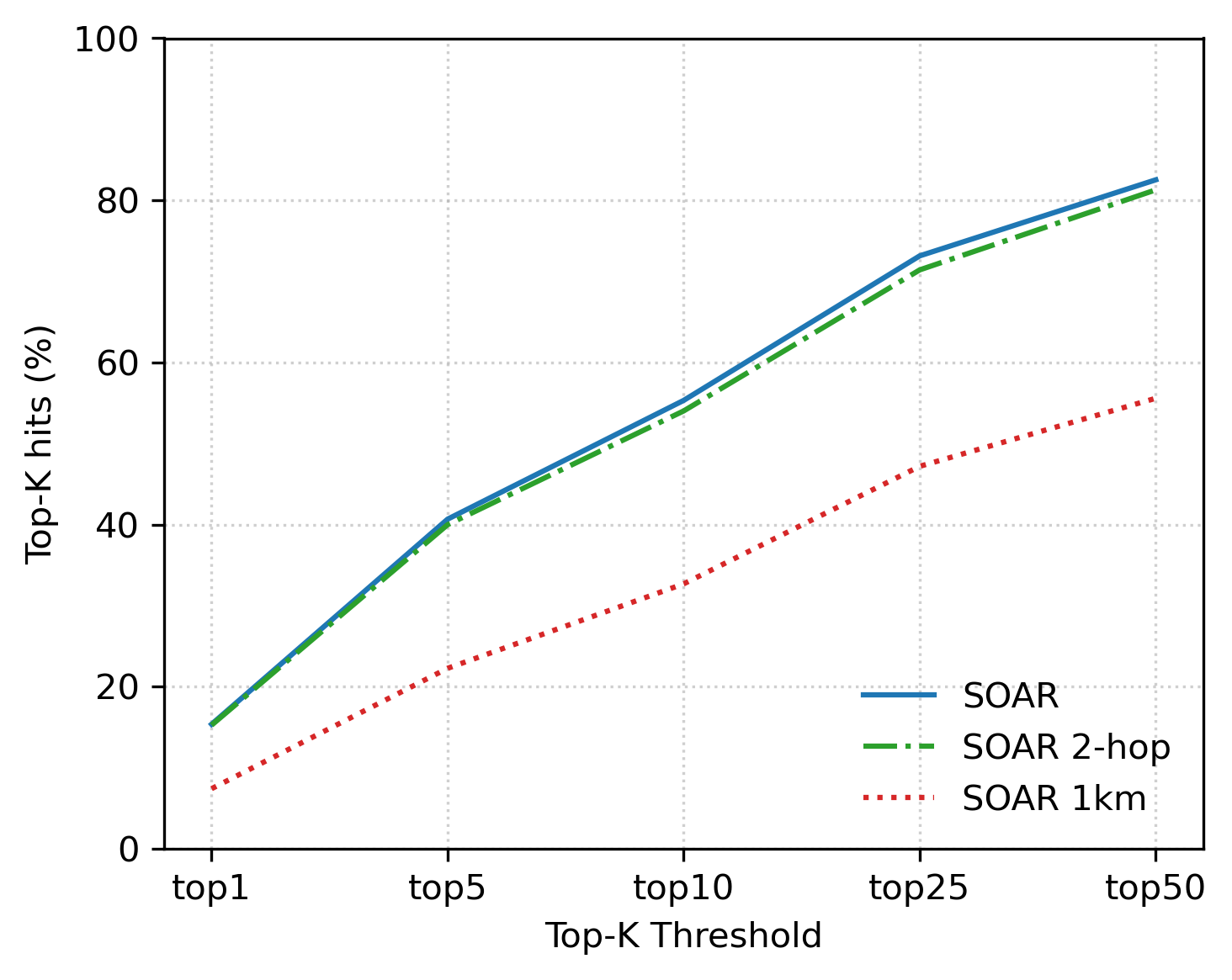}
        \caption{Top-K Hits}
        \label{fig:topk_hits_3models}
    \end{figure}

We compared the performance of our model on the corresponding potential locations between ranges of 400 meters and 1 km as illustrated in Figures~\ref{fig:accuracy_3models} through \ref{fig:topk_hits_3models} . We compared them in terms of accuracy in Figure~\ref{fig:accuracy_3models}, recall in Figure~\ref{fig:recall_3models}, F1-score in Figure~\ref{fig:f1_3models}, precision in Figure~\ref{fig:Precision_3models} and top-K hits in Figure~\ref{fig:topk_hits_3models}. Table~\ref{tab:model_versions} shows the differences in Recall, F1-score and accuracy between the different versions of our model. As the number of potential locations increases exponentially when range increases, we observe the expected drop in related metrics. We chose to move forward with the decreased range as users' preference to visit a location further away from their present location is increasingly lower.

Another important factor, especially for the Top-1 and Top-5 categories, we observe that the accuracy is significantly greater than the top-K hits in general. This indicates that the users' most preferential location choice is heavily affected by their friends' reviews and mentality rather than the popular opinion.



\begin{table}[htbp]
\centering
\setlength{\tabcolsep}{4pt} 
\renewcommand{\arraystretch}{1.2} 
\caption{Comparison of Recall, F1-score, and Accuracy among versions of our model 2-hop, and 1km across different top-K categories.}
\label{tab:model_versions}
\begin{tabular}{lccccccccc}
\toprule
\multirow{2}{*}{\textbf{Top-K}} & \multicolumn{3}{c}{\textbf{Recall (\%)}} & \multicolumn{3}{c}{\textbf{F1-score (\%)}} & \multicolumn{3}{c}{\textbf{Accuracy (\%)}} \\
\cmidrule(lr){2-4} \cmidrule(lr){5-7} \cmidrule(lr){8-10}
 & \textbf{Main} & 2-hop & 1km & \textbf{Main} & 2-hop & 1km & \textbf{Main} & 2-hop & 1km \\
\midrule
Top-1  & \textbf{16.37} & 16.25 & 7.98  & \textbf{27.17} & 27.00 & 14.52 & \textbf{34.22} & 34.13 & 29.58 \\
Top-5  & \textbf{42.22} & 41.68 & 23.56 & \textbf{54.72} & 54.34 & 36.32 & \textbf{47.66} & 47.53 & 38.11 \\
Top-10 & \textbf{56.59} & 55.62 & 34.25 & \textbf{65.08} & 64.61 & 47.68 & \textbf{54.54} & 54.39 & 43.72 \\
Top-25 & \textbf{73.66} & 72.45 & 48.67 & \textbf{74.53} & 74.18 & 59.72 & \textbf{62.29} & 62.21 & 50.81 \\
Top-50 & \textbf{82.76} & 81.83 & 56.77 & \textbf{79.05} & 78.78 & 65.37 & \textbf{66.89} & 66.73 & 54.63 \\
\bottomrule
\end{tabular}
\end{table}

We also tested the performance of our algorithm where we include the friends of friends in the recommendation process. This exponentially increases execution time, more than 8 times slower, as shown in Table~\ref{tab:speed_comparison} resulting in a non-viable time especially in large datasets. Additionally, as also seen on Table~\ref{tab:model_versions} the differences between our main algorithm and the 2-hop version is marginally different in the end result, if not slightly worse in some cases. This indicates that the user's friends are enough to determine the user's tendencies and including friends of friends may dilute the collection of traits important to the user.

\begin{table}[!ht]
\centering
\caption{Performance comparison between friends and friends of friends}
\label{tab:speed_comparison}
\begin{tabular}{lcccc}
\toprule
\textbf{Model} & \textbf{Speed (predictions/s)} & \textbf{Total Execution Time} \\ 
\midrule
SOAR 2 HOP & 7.32 & 36 hours \\
SOAR & 55.56 & 4.6 hours \\ 
\bottomrule
\end{tabular}
\end{table}

\begin{figure}
        \centering
        \includegraphics[width=0.5\textwidth]{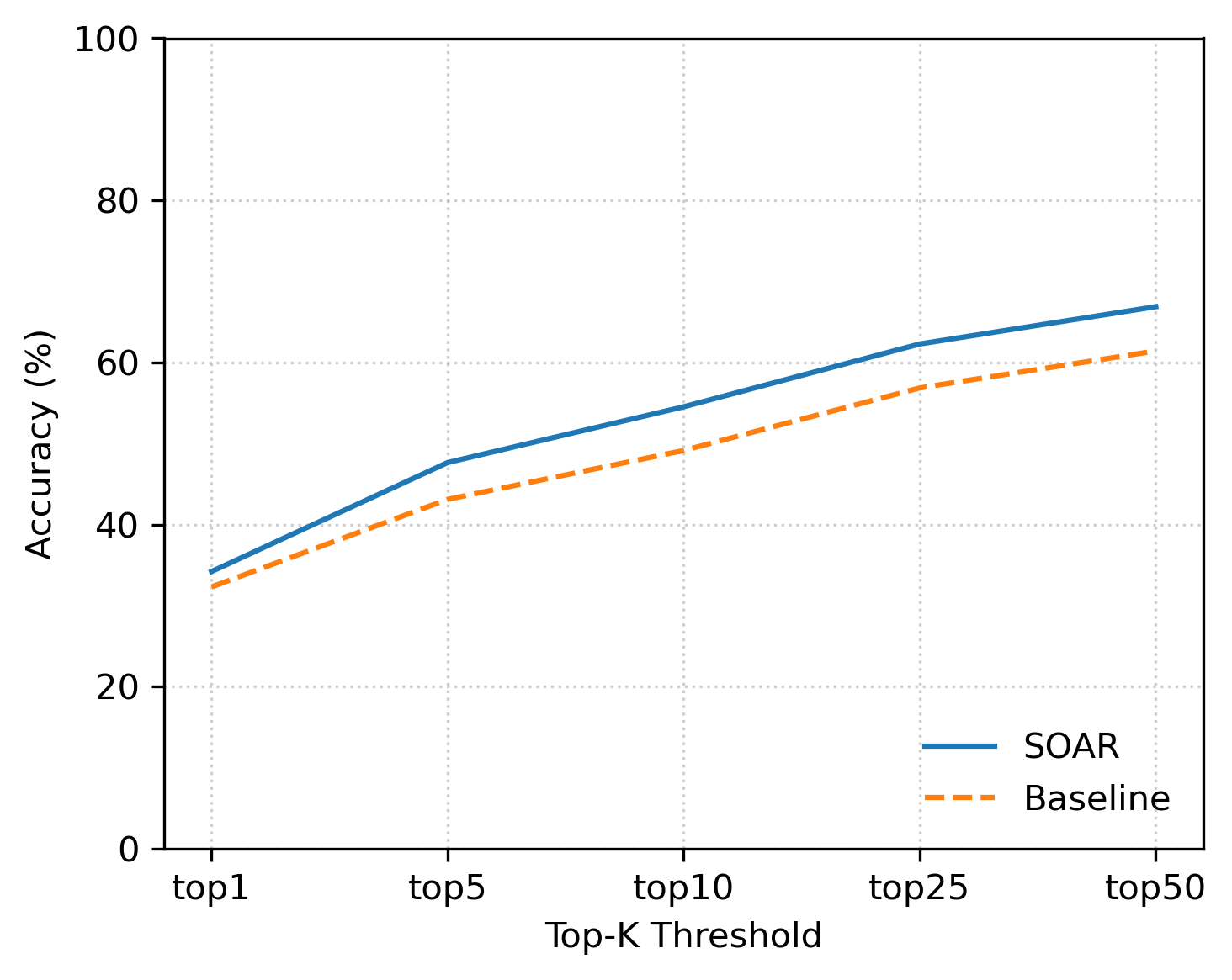}
        \caption{Accuracy comparison between models}
        \label{fig:accuracy_comparison}
    \end{figure}

\begin{figure}
        \centering
        \includegraphics[width=0.5\textwidth]{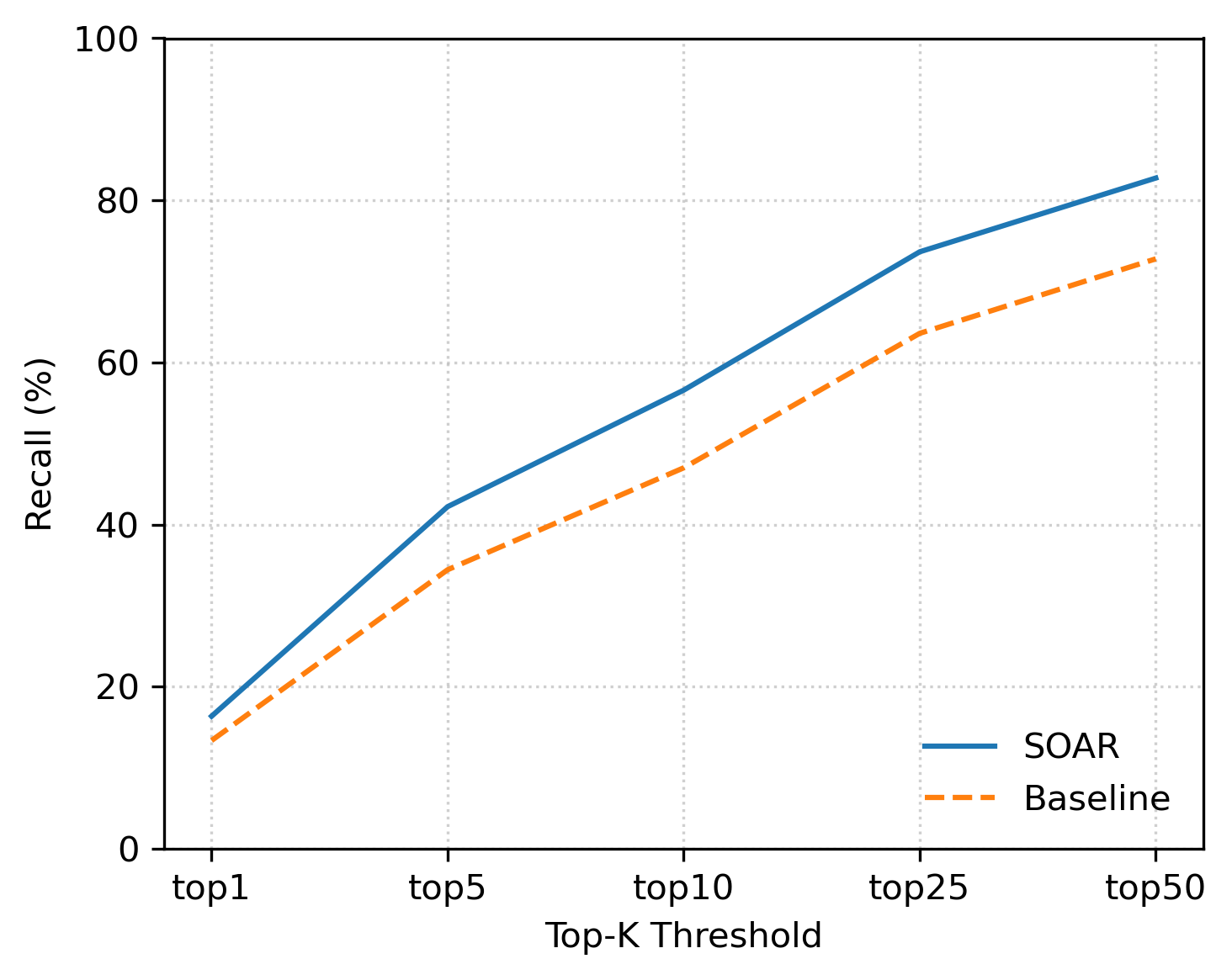}
        \caption{Recall comparison between models}
        \label{fig:recall_comparison}
    \end{figure}

\begin{figure}
        \centering
        \includegraphics[width=0.5\textwidth]{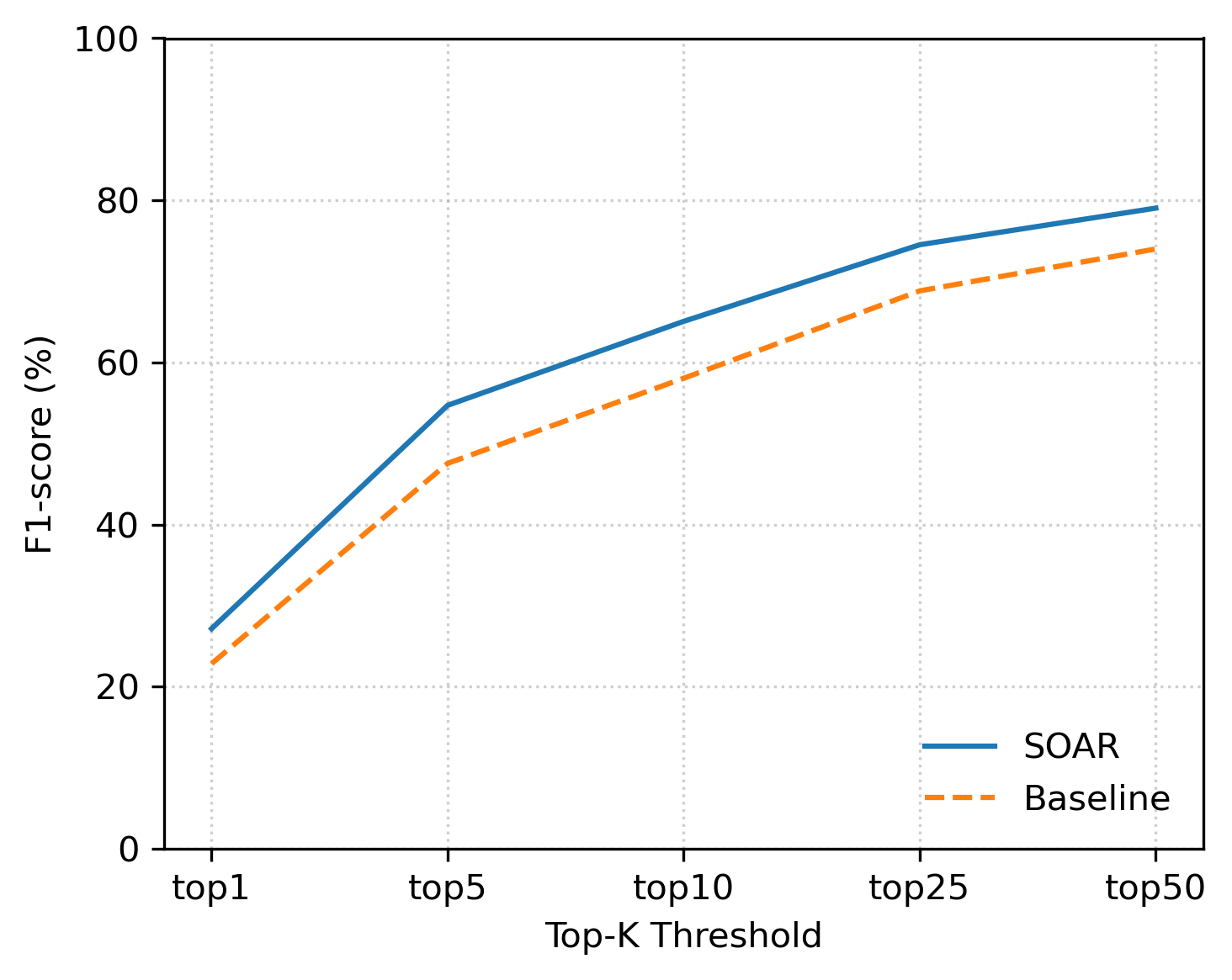}
        \caption{F1-score comparison between models}
        \label{fig:f1_comparison}
    \end{figure}

\begin{figure}
        \centering
        \includegraphics[width=0.5\textwidth]{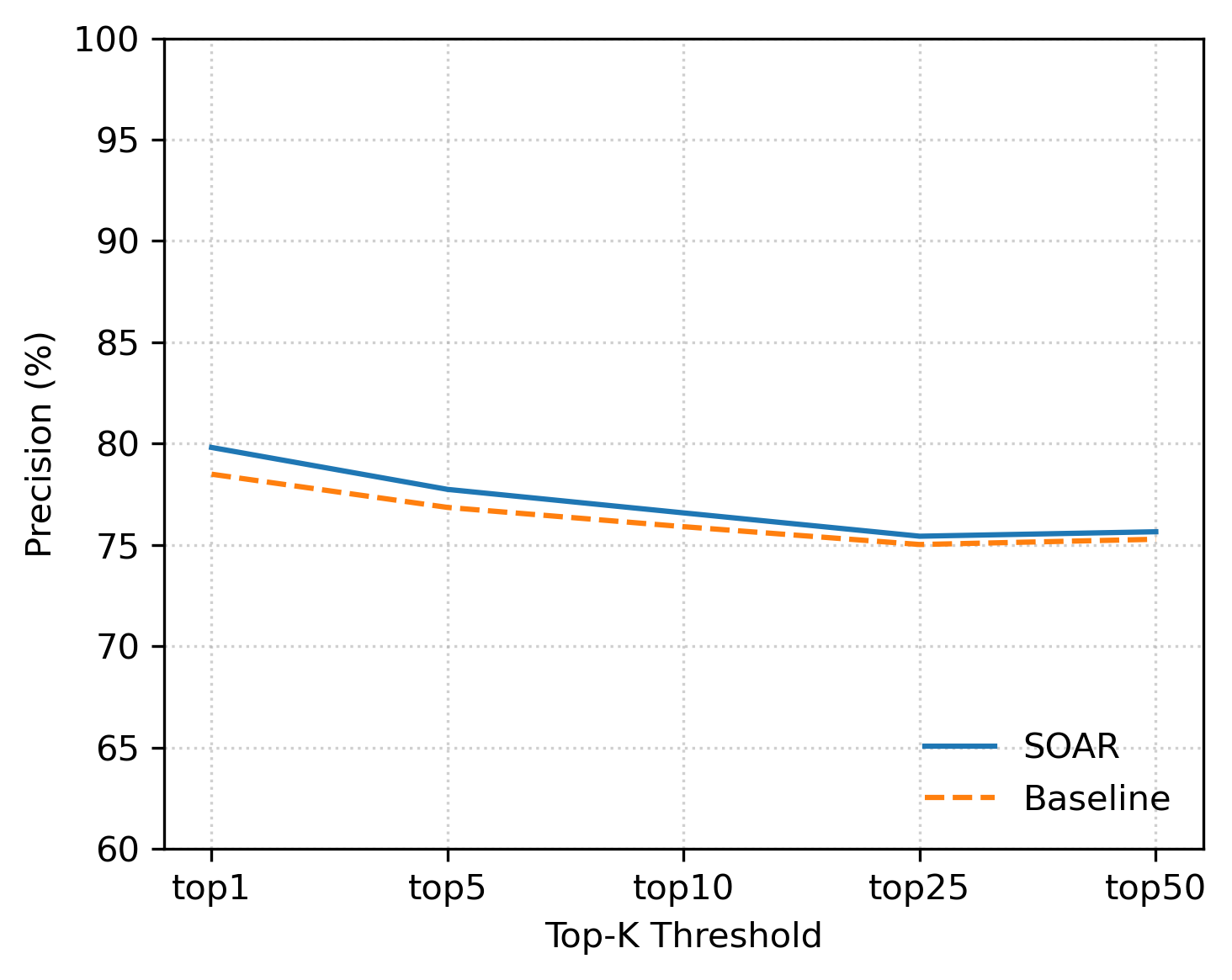}
        \caption{Precision comparison between models}
        \label{fig:precision_comparison}
    \end{figure}

\begin{figure}
        \centering
        \includegraphics[width=0.5\textwidth]{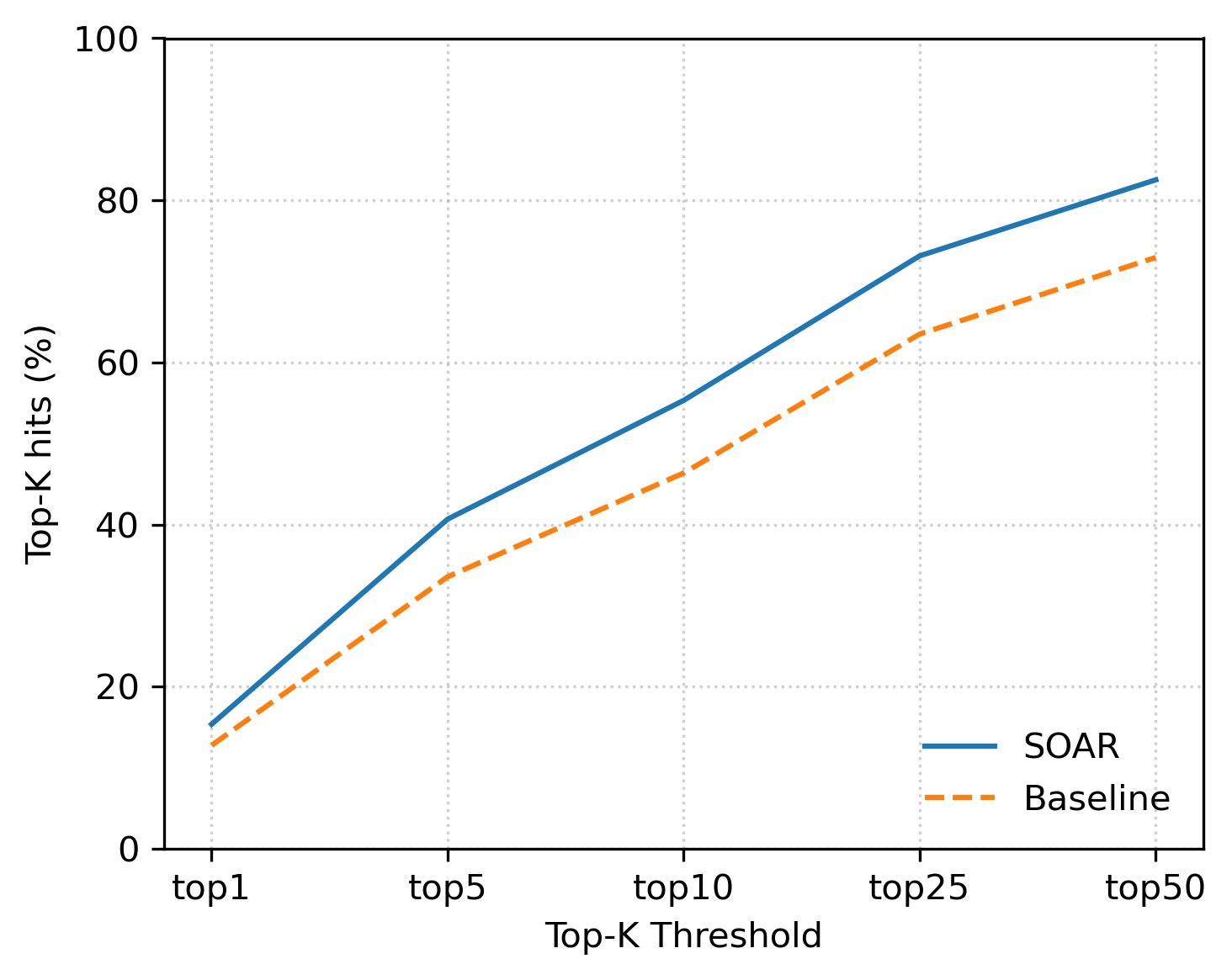}
        \caption{Top-K Hits comparison between models}
        \label{fig:Topkhits}
    \end{figure}

We compared the proposed model against a baseline method that does not combine textual reviews and the friendship network of the users. Figures \ref{fig:accuracy_comparison} to \ref{fig:Topkhits} illustrate the improvement in recommendation performance across all evaluated metrics described in the previous section: accuracy in Figure~\ref{fig:accuracy_comparison}, recall in Figure~\ref{fig:recall_comparison}, f1-score in Figure~\ref{fig:f1_comparison}, precision in Figure~\ref{fig:precision_comparison} and top-K hits in \ref{fig:Topkhits}. The comparison spans multiple Top-K categories, ranging from Top-1 to Top-50, demonstrating the performance achieved by the proposed model. The differences can also be seen at Table~\ref{tab:model_comparison}.

\begin{table}[htbp]
\centering
\caption{Comparison of SOAR and baseline on Recall, F1, and top-K Hits}
\begin{tabular}{lcccccc}
\hline
\multirow{2}{*}{\textbf{Top-K}} & 
\multicolumn{2}{c}{\textbf{Recall (\%)}} & 
\multicolumn{2}{c}{\textbf{F1 (\%)}} & 
\multicolumn{2}{c}{\textbf{Top-K Hits (\%)}} \\

 & SOAR & baseline & SOAR & baseline & SOAR & baseline \\
\hline
Top-1  & \textbf{16.37} & 13.37 & \textbf{27.17} & 22.85 & \textbf{15.38} & 12.77 \\
Top-5  & \textbf{42.22} & 34.44 & \textbf{54.72} & 47.56 & \textbf{40.68} & 33.57 \\
Top-10 & \textbf{56.59} & 47.01 & \textbf{65.08} & 58.06 & \textbf{55.33} & 46.37 \\
Top-25 & \textbf{73.66} & 63.60 & \textbf{74.53} & 68.84 & \textbf{73.16} & 63.51 \\
Top-50 & \textbf{82.76} & 72.79 & \textbf{79.05} & 74.01 & \textbf{82.54} & 72.96 \\
\hline
\end{tabular}
\label{tab:model_comparison}
\end{table}



We observe a notable increase in performance, especially in the top-5 and top-10 categories which are practically the first page in a recommendation application. The increased accuracy indicates the effect of user's reviews and the tendencies of the user to follow their friend's habits and share common preferences with them.

We also observe that precision is decreasing while recall, F1, and accuracy increase with larger K-stems from the inherent trade-off between prediction strictness and coverage in ranked recommendation. When the cutoff K expands, the model produces more predictions - positive classifications, capturing a higher proportion of liked POIs, increasing true positives and recall. However, this broader inclusion also introduces more unliked locations among the top-K - increasing false positives, reducing the proportion of correct recommendations and thus lowering precision. In a real-world application, this phenomenon can be explained by the fact that as one moves further down the recommendation list, the likelihood of encountering locations the user does not like increases.

We also evaluated our model using the top-K Hits metric, as illustrated in Figure~\ref{fig:Topkhits}. The top-K Hits measure provides an overall indication of the model’s recommendation accuracy within the top K predicted POIs, irrespective of the user’s eventual sentiment. More specifically, a hit occurs when a POI recommended by the system is actually visited by the user, while a miss indicates that the user visited a POI that was not among the top-K recommendations. This metric captures how effectively the model can anticipate user behavior based solely on ranking relevance, offering an objective measure of recommendation performance that complements sentiment-based evaluations.

\section{Conclusions}
\label{sec:conclusions}

Point-of-Interest (POI) and item recommendation tasks present significant challenges in the modern era due to the multitude of factors that influence user decision-making. Conventional approaches often capture only a subset of these factors or fail to effectively integrate them. However, since user preferences emerge from the interaction of many interdependent elements rather than isolated influences, it is crucial not only to identify these factors but also to understand their relationships and combined impact. The most challenging aspect of recommendation remains achieving genuine personalization for each individual user.

In this work, we introduced a novel framework that extracts and links meaningful aspects from review texts to uncover individual user tendencies. By incorporating both public sentiment and social influence derived from users’ friends, our approach provides personalized and context-aware POI recommendations. Experimental results demonstrate that integrating these textual and social factors significantly enhances recommendation performance.

Future research could explore the integration of additional contextual signals such as temporal patterns and mobility data. Moreover, applying more advanced language models or graph-based learning methods may further enhance the system’s ability to capture nuanced user preferences and dynamic relationships among locations.

\bibliography{bibliography}

\end{document}